\pdfoutput=1

\documentclass[11pt]{article}

\usepackage{emnlp2021}

\usepackage{times}
\usepackage{amsmath}
\usepackage{amssymb}
\usepackage{latexsym}
\usepackage{multicol}
\usepackage{multirow}
\usepackage{diagbox}
\usepackage{booktabs}
\usepackage{subcaption}
\usepackage{graphicx}
\usepackage{graphics}
\usepackage{stfloats}
\usepackage[ruled, linesnumbered]{algorithm2e}
\usepackage{float}
\usepackage{array}

\usepackage[T1]{fontenc}

\usepackage[utf8]{inputenc}

\usepackage{microtype}

\title{MuVER: Improving First-Stage Entity Retrieval with \\ Multi-View  Entity Representations}

\author{
Xinyin Ma$^{\diamond\ddagger}$, Yong Jiang$^{\dagger}$\textsuperscript{$\ast$}, Nguyen Bach$^{\dagger}$, Tao Wang$^{\dagger}$, \\  \textbf{Zhongqiang Huang}$^{\dagger}$, \textbf{Fei Huang$^{\dagger}$, Weiming Lu$^{\diamond}$}\thanks{\hspace{1mm} Corresponding authors.\newline \hspace*{1.5em} $^{\ddagger}$ Work was done when Xinyin Ma was interning at Ali-baba DAMO Academy.} \\
$^{\diamond}$ College of Computer Science and Technology , Zhejiang University \\
$^{\dagger}$ DAMO Academy, Alibaba Group \\
{\tt \{maxinyin, luwm\}@zju.edu.cn, yongjiang.jy@alibaba-inc.com}
}
\begin{document}
\maketitle
\begin{abstract}
  Entity retrieval, which aims at disambiguating mentions to canonical entities from massive KBs, is essential for many tasks in natural language processing. Recent progress in entity retrieval shows that the dual-encoder structure is a powerful and efficient framework to nominate candidates if entities are only identified by descriptions. However, they ignore the property that meanings of entity mentions diverge in different contexts and are related to various portions of descriptions, which are treated equally in previous works.
  In this work, we propose \textbf{Mu}lti-\textbf{V}iew \textbf{E}ntity \textbf{R}epresentations (MuVER), a novel approach for entity retrieval that constructs multi-view representations for entity descriptions and approximates the optimal view for mentions via a heuristic searching method. Our method achieves the state-of-the-art performance on ZESHEL and improves the quality of candidates on three standard Entity Linking datasets\footnote{Our source code is available at \url{https://github.com/Alibaba-NLP/MuVER}.}. 

\end{abstract}

\section{Introduction}
Entity linking (EL) refers to the task that disambiguates the mentions in textual input and retrieves the corresponding unique entity in large Knowledge Bases (KBs) \citep{HanSZ11, CeccarelliLOPT13}. The majority of neural entity retrieval approaches consist of two steps: Candidate Generation \citep{PershinaHG15, Zwicklbauer2016}, which nominates a small list of candidates from millions of entities with low-latency algorithms, and Entity Ranking \citep{YangIR18, LeT19, DeCao2020}, which ranks those candidates to select the best match with more sophisticated algorithms. 

In this paper, we focus on the Candidate Generation problem (a.k.a. the first-stage retrieval). Prior works filter entities by alias tables \citep{Fang2019Reinforce} or precalculated mention-entity prior probabilities, e.g., $p(entity|mention)$ \citep{PhongLe18a}. \citet{Ganea2017Deep} and \citet{Yamada2016Joint} build entity embedding from the local context of hyperlinks in entity pages or entity-entity co-occurrences. Those embedding-based methods were extended by BLINK \citep{WuPJRZ20} and DEER \citep{Daniel2019DenseRetrieval} to two-tower dual-encoders \citep{ Khattab2020Colbert}, which encode mentions and descriptions of entities into high-dimensional vectors respectively. Candidates are retrieved by nearest neighbor search \citep{AndoniI08, FAISS} for a given mention. Solutions that require only entity descriptions \citep{LogeswaranCLTDL19} are scalable, as descriptions are more readily obtainable than statistical or manually annotated resources. 

\begin{figure*}[t]
   \begin{center}
   \scalebox{1}{
      \includegraphics[width=1.\linewidth]{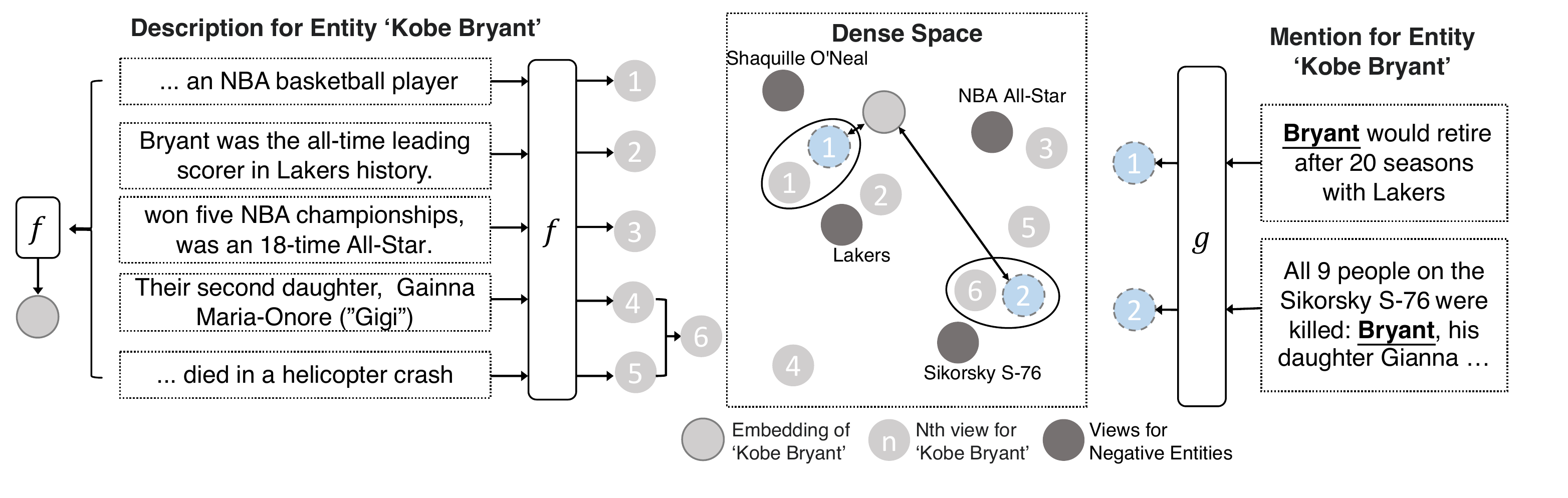}
    }
   \end{center}
   \vspace{-4mm}
      \caption{An illustration of our MuVER framework. (i) The contextual information of the given document for the same mention may differ widely (Right), resulting in a large discrepancy between their representations (Blue circles with dashed borders) and the embedding of ``Kobe Bryant'' has trouble in getting close to both of them. (ii) We refer to each sentence as a view for descriptions to form a view set $V$ (Gray circles with number) and merge views to approximate the optimal views for mentions (points enclosed by ellipses). 
      } \label{fig:introduction}
\end{figure*}

Although description-based dual-encoders can compensate for the weakness of traditional methods and have better generalization ability to unseen domains, they aim to map mentions with divergent context to the same high-dimensional entity embedding. As shown in Figure \ref{fig:introduction}, the description of ``Kobe Bryant'' mainly concentrates on his professional journey. As a result, the embedding of ``Kobe Bryant'' is close to the context which describes the career of Kobe but is semantically distant from his helicopter accident. Dual-encoders are trained to encode those semantically divergent contexts to representations that are close to the embedding of ``Kobe Bryant''. The evidence relies on the Figure \ref{fig:effect_length} (section \ref{exp:wika}) that the previous method \citep{WuPJRZ20} is good at managing entities with short descriptions but seems troubling to retrieve entities with long descriptions, which contains too much information to be encoded into a single fixed-size vector.

 


To tackle those issues, we propose to construct multi-view representations from descriptions. The contributions of our paper are as follows:  

\begin{itemize}
    \item We propose an effective approach, MuVER, for first-stage entity retrieval, which models entity descriptions in a multi-view paradigm.
    \item We define a novel distance metric for retrieval, which is established upon the optimal view of each entity. Furthermore, we introduce a heuristic search method to approximate the optimal view.
    \item MuVER achieves state-of-the-art performance on ZESHEL and generates higher-quality candidates on AIDA-B, MSNBC and WNED-WIKI in full Wikipedia settings. 
\end{itemize}

\section{Methods}
\subsection{Problem Setup}
Formally, given an unstructured text $D$ with a recognized mention $m$, the goal of entity linking is to learn a mapping from the mention $m$ to the entity entry $e$ in a knowledge base $\mathcal{E} = \{e_1, e_2, \ldots, e_N\}$, where $N$ can be extremely large (for Wikipedia, $N=5.9M$). In the literature, existing  retrieval methods address this problem in a two-stage paradigm: (i) selecting the top relevant entities to form a candidate set $\mathcal{C}$
where $|\mathcal{C}| \ll |\mathcal{E}|$; (ii) ranking candidates to find the best entity within $\mathcal{C}$. In this work, we mainly focus on the first-stage retrieval, following \citet{LogeswaranCLTDL19}'s setting to assume that for each $e \in \mathcal{E}$, entity title $t$ and description $d$ are provided in pairs. 


\subsection{Multi-View Entity Representations}

\paragraph{Dual-encoders} We tackle entity retrieval as a matching problem, where two separated encoders, entity encoder $f$ and mention encoder $g$, are deployed. We consider BERT \citep{Devlin2019BERT} as the architecture to encode textual input, which can be formulated as:

    \begin{align*} 
          & f(t, d) = T_{1}([CLS] \ t \ [ENT] \ d \ [SEP]) \\
          & g(m) = T_{2}([CLS] \  \text{ctx}_l \ [M_s] \ m \ [M_e] \ \text{ctx}_r \ [SEP])
    \end{align*}

where $t$, $d$, $m$, $\text{ctx}_l$, $\text{ctx}_r$ refer to word-pieces tokens of the entity title, the entity description, the mention and the context before and after the mention correspondingly. Besides, we use $[M_s]$ and $[M_e]$ to denote the {\textit{start \ of \ mention}} and {\textit{end \ of \ mention}} identifiers respectively. The special token $[ENT]$ serves as the delimiter of titles and descriptions. $T_{1}$ and $T_{2}$ are two independent BERT, with which we estimate the similarity between mention $m$ and entity $e$ as $sim(m, e) = f(t, d) \cdot g(m)$.

\paragraph{Multi-view Description} 

Our method matches a mention to the appropriate entity by comparing it with entity descriptions. Motivated by the fact that mentions with different contexts correspond to different parts in descriptions, we propose to construct multi-view representations for each description. Specifically, we segment a description into several sentences. We refer to each sentence as a view $v$, which contains partial information, to form a view set $\mathcal{V}$ of the entity $e$. Figure \ref{fig:introduction} illustrates an example that constructs a view set $\mathcal{V}$ for ``Kobe Bryant''. 


\paragraph{Multi-view Matching} 
Given a view set $\mathcal{V}=\{v_1, v_2, \ldots, v_k\}$ for entity $e$, determining whether a mention $m$ matches the entity $e$ requires a metric space to estimate the relation between $m$ and $\mathcal{V}$, which can be defined as
\begin{align}
d(m, \mathcal{V}) = \|g(m)-f(t, [v_1, v_2, ..., v_k, v_i \in \mathcal{V}])\|
\end{align}
where $[v_1, v_2, ..., v_k]$ refers to an operation that concatenates tokens in views following the sentence order in descriptions and $t$ is the corresponding entity title for $\mathcal{V}$. Note that this metric can be applied to the subset of $\mathcal{V}$ to focus on partial information of the description. As mentioned before, for $m$ in different contexts, only a part of the views are related. For each mention-entity pair $(m, e)$ and the view set $\mathcal{V}$ of $e$, we define its optimal $Q^{*}$ as:
\begin{align}
Q^{*}(m, e) \triangleq {\arg\min}_{ Q \subseteq \mathcal{V} } d(m, Q) 
\label{eqn:optimal_view_set}
\end{align}
where $Q$ is a subset of $\mathcal{V}$ and $Q^{*}$ has the minimal distance to current mention $m$. We define the distance $d(m, Q^{*}(m, e))$ as the matching distance between $e$ and $m$. To find the optimal entity for mention $m$, we select the entity that has minimal matching distance:
\begin{align}
e^{*} = {\arg\min}_{ e \in \{e_1, e_2, \ldots, e_N\} } d(m, Q^{*}(m, e)) \label{eqn:retrieval}
\end{align}


\paragraph{Distance Metric \& Training Objectives}
The above retrieval process requires an appropriate metric space to estimate the similarity between views and mentions. The metric space should satisfy that similar inputs are pulled together and dissimilar ones are pushed apart. To achieve this, we introduce an NCE loss \citep{Aaron2018CPC} to establish the metric space : 
    \begin{align*}
        \mathcal{L}_{NCE} = 
            \underset{\mathcal{E}^{'}}{\mathop{\mathbb{E}}}\left[
                     \log \frac{\exp(d(m, Q^{*}(m, e)))}
                     {\sum_{e_i \in \mathcal{E}^{'}} \exp(d(m, Q^{*}(m, e_i)))} \right]
    \end{align*} 
where $\mathcal{E}^{'} = \{e\} \cup \{e_1, \ldots, e_{n-1}\}$. Mention-entity pairs $(m, e)$ are pulled together and randomly sampled $n-1$ negatives $ \{e_1, \ldots, e_{n-1}\}$ are pushed apart from $m$, based on their matching distance in the current metric space.  Unfortunately, $Q^{*}(m, e)$ is intractable due to the non-differentiable subset operation in Equation \ref{eqn:optimal_view_set}. Besides, it is time-consuming to obtain the optimal view by checking all subsets exhaustively. In this work, we consider a subset that contains only one view to approximate it. Specifically, we select the best $v^*(m, e)\triangleq\arg\min_{v\in \mathcal{V}}d(m, \{v\})$ from $\mathcal{V}$ as an alternative to the optimal view $Q^*$:
\begin{align}
    d(m, Q^{*}(m, e))  \approx d(m, v^*(m, e))) \label{eqn:approximation}
\end{align} 
Note that this approximation can be done in time complexity of $O(N)$, which simply selects a view with minimal distance to the given mention. Using Equation \ref{eqn:approximation}, we can rewrite the NCE loss as:
\resizebox{\linewidth}{!}{
\begin{minipage}{\linewidth}
    \begin{align*}
        \mathcal{L}_{NCE} = 
            \underset{\mathcal{E}^{'}}{\mathop{\mathbb{E}}}\left[
                     \log \frac{\exp(d(m, \{v^{*}(m, e)\}))}
                     {\sum_{e_i \in \mathcal{E}^{'}} \exp(d(m, \{v^{*}(m, e_i)\}))} \right]
    \end{align*} 
\end{minipage}
}

\subsection{Heuristic Searching for Inference}
The approximation in Equation \ref{eqn:approximation} obviously can not reveal the matching distance because $v^*(m, e)$ contains insufficient information for retrieval. We want to search for a better view $Q^{'} \subset \mathcal{V}$ that $d(m, Q^{'}) < d(m, v^*(m, e))$. 

Combining views $(Q_1, Q_2)$ that contain complementary information is more likely to incorporate richer information into the newly assembled view. 
Considering two sets $Q_1\subset\mathcal{V}$ and $Q_2\subset\mathcal{V}$ and a distance metric $d(Q1, Q2)=\|f(t, Q_1) -f(t, Q_2)\|$, where $t$ is the title of the entity and $f$ represents the entity encoder, the most distant pair of views $(Q_1, Q_2)$ achieve the largest magnitude on $d(Q_1, Q_2)$ among all pairs and is interpreted as the pair of views with less shared information.
For each iteration, We search the top-k distant pairs $(Q_1, Q_2)$ to form a new view $Q^{'} = Q_1 \cup Q_2$ and expand $Q^{'}$ into $\mathcal{\mathcal{V}}$ to encode the merged $Q'$ by $f(t, Q^{'})$ to produce a new representation for the involved entity.
Searching and merging are performed iteratively until $|\mathcal{V}|$ reaches the maximal allowable value or the number of iterations reaches the preset value. During the inference, we precompute and cache the representations of views and select the view with minimal distance to $m$.

\section{Experiments}

\definecolor{mygreen}{rgb}{0.1,0.8,0.1}
\def\scoreup#1{$(\color{mygreen} \uparrow #1)$}
\def\scoredown#1{$(\color{red} \downarrow #1$)}

\begin{table*}[t]
   \centering
   \scalebox{0.9}{
   \begin{tabular}{c |c c c c c c c c}
      \midrule
      \midrule
       \bf{Method}  & \bf R@1  & \bf R@2 & \bf R@4 & \bf R@8 & \bf R@16 & \bf R@32 & \bf R@50 & \bf R@64 \\
      \midrule
      BM25 & - & - &  - & - & - & - & - &  69.13 \\
      BLINK\citep{WuPJRZ20} & - & - & - & - & - & - & - & 82.06 \\
      \citet{Partalidou2021improving} & - & - & - & - & - & - & 84.28 & - \\
      BLINK $ \text{\citep{WuPJRZ20}}^\dagger$ & \bf{46.51} & 58.22 & 67.00 & 72.77 & 77.29 & 81.03 & 83.38 & 84.78  \\
      BLINK $ \text{\citep{WuPJRZ20}}^* $ & 45.59 & 57.55 & 66.10 & 72.47 & 77.65 & 81.69 & 84.31 & 85.56  \\
      SOM \citep{Zhang2021under} & - & - & - & - & - & - & - & 87.62 \\
      \midrule
      MuVER (w/o Heuristic Search) & 43.49 & 58.56 & 68.78 & 75.87 & 81.33 & 85.86 & 88.35 & 89.52\\
      MuVER  & 45.40 & \bf{60.84} & \bf{71.26} & \bf{78.27} & \bf{83.19} & \bf{87.58} & \bf{89.75} & \bf{90.84}  \\
      \midrule
   \end{tabular} 
   }
   \vspace{-1mm}
   \caption{Recall@k (R@k) on the test set of ZESHEL to retrieve entities from Wikia. $\dagger$We reproduce BLINK and achieve a higher result compared with the result reported in the paper. * expands context length to 512. For SOM, we report the performance using in-batch negatives to have a fair comparison. } \label{tbl:zeshel_main}
   \vspace{-2mm}
\end{table*} 

\begin{table*}[t]
   \centering
   \begin{tabular}{c |c c c |c c c |c c c}
      \midrule
      \midrule
       & \multicolumn{3}{c|}{AIDA-b} & \multicolumn{3}{c|}{MSNBC} & \multicolumn{3}{c}{WNED-WIKI}\\
       & R@10 & R@30 & R@100 & R@10 & R@30 & R@100   & R@10 & R@30 & R@100\\
       \midrule
       BLINK  &92.38 & 94.87 & 96.63 & 93.03 & 95.46 & 96.76 & 93.47 & 95.46 & \bf{97.76} \\
       MuVER  & \bf{94.53}  & \bf{95.25} & \bf{98.11} & \bf{95.02} & \bf{96.62} & \bf{97.75} & \bf{94.05} & \bf{95.78} & 97.34\\
      \midrule
   \end{tabular} 
   \caption{Results on three standard Entity Linking datasets. We test our model under the setting that only descriptions of entities are available. The number of basic views for each entity is 5.}
   \vspace{-2mm} \label{tbl:wiki}
\end{table*} 

\subsection{Datasets}
We evaluate MuVER under two different knowledge bases: Wikia, which the Zero-shot EL dataset is built upon, and Wikipedia, which contains 5.9M entities. We select one in-domain dataset, AIDA-CoNLL \citep{AIDAConll}, and two out-of-domain datasets, WNED-WIKI \citep{DBLP:journals/semweb/GuoB18} and MSNBC \citep{MSNBC}, from standard EL datasets to validate MuVER in the full Wikipedia setting. Statistics of datasets are listed in Appendix \ref{apx:statistic}.

\subsection{KB: Wikia} \label{exp:wika}
\citet{LogeswaranCLTDL19} constructs a zero-shot entity linking dataset (ZESHEL), which places more emphasis on understanding the unstructured descriptions of entities to resolve the ambiguity of mentions on four unseen domains. 
 
Concretely, MuVER uses BERT-base for $f$ and $g$ to make a fair comparison with previous works. We adopt an adam optimizer with a small learning rate $1\text{e}^{-5}$ and $10\%$ warmup steps. We use batched random negatives and set the batch size to 128. The max number of context tokens is 128 and the max number of view tokens equals 40. Training 20 epochs takes one hour on 8 Tesla-v100 GPUs. 

We compare MuVER with previous baselines in Table \ref{tbl:zeshel_main}. Since MuVER is not limited by the length of descriptions, we add another baseline to extend BLINK to have 512 tokens (which is the max number of tokens for BERT-base). As shown in the table, we exceed BLINK by 5.28\% and outperform SOM by 3.22\% on Recall@64. We observe that Recall@1 of MuVER is lower than BLINK and the heuristic searching method can alleviate this problem. Detailed results on unseen domains are listed in Appendix \ref{apx:unseen}. 

\paragraph{Effect of Heuristic Search}
We compare two distance-based merging strategies: taking closer or farther pairs of views to merge. We find out that merging views whose sentences are adjacent to each other in the original unstructured descriptions is a computationally efficient way to select the combined views. Table \ref{tbl:combine_MI} shows that as the number of views increases, MuVER yields higher-quality candidates while the opposite strategy is troubled to provide more valuable views.
Besides, our method can be regarded as a generalized form of SOM \citep{Zhang2021under} and BLINK \citep{WuPJRZ20}, which contain 128 views and one view correspondingly. SOM computes the similarity between mentions and tokens in descriptions, which stores 128 embeddings for each entity. Compared with SOM, MuVER reduces the number of views to a smaller size with improved quality, which is more efficient and effective. 

\begin{table}[h]
   \centering
   \scalebox{0.85}{%
       \begin{tabular}{c | c | c | c l}
          \midrule
          \midrule
            \multicolumn{4}{c}{\bf{Without View Merging}} \\
            \midrule
            Methods & \# of Views & \multicolumn{2}{c}{Recall@64} \\
            \midrule
            BLINK & 1 & \multicolumn{2}{c}{85.56} \\
            SOM & 128  & \multicolumn{2}{c}{87.62} \\
            MuVER& 15.33&  \multicolumn{2}{c}{89.52}  \\
            \midrule
            \multicolumn{4}{c}{\bf{With View Merging}} & \\ [2pt]
            \midrule
            Methods & \# of Views & Distant Pairs& Close Pairs \\
            \midrule
            \multirow{5}{*}{MuVER}
            & 21.07& 90.15 & 89.95  \\
            & 26.18& 90.51 & 89.99  \\
            & 28.39& 90.66 & 89.98  \\
            & 30.48& 90.79 & 89.89  \\
            & 32.48& 90.84 & 89.92  \\
          \midrule
       \end{tabular} %
    }
       \vspace{-2mm}
       \caption{Recall@64 on ZESHEL with varying number of views. We shot different merging strategies and ``Distant Pairs'' refers to our Heuristic Search method.}
       \label{tbl:combine_MI}
\end{table} 

\begin{figure}[t]
   \begin{center}
   \scalebox{1}{
      \includegraphics[width=0.95\linewidth]{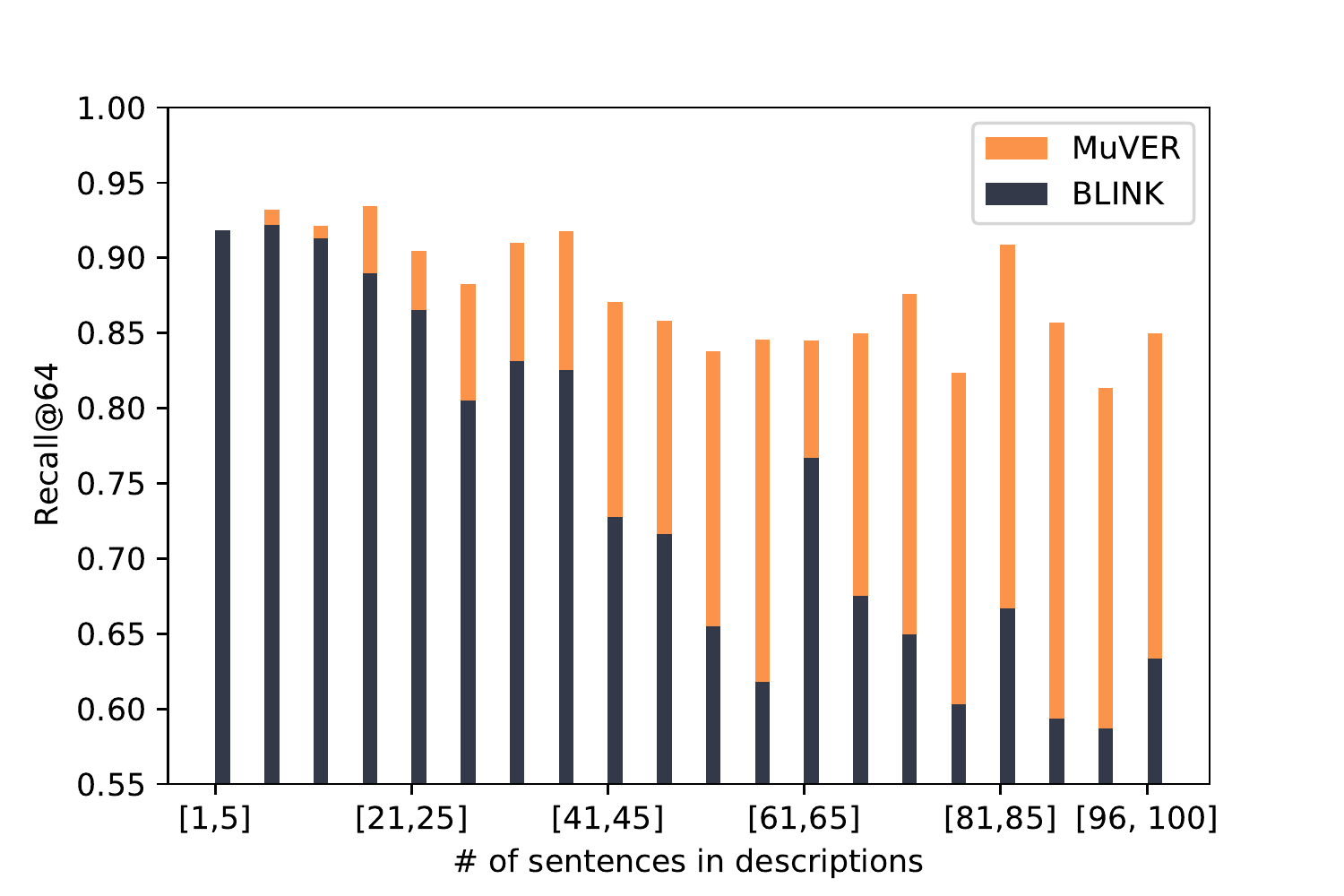}
    }
   \end{center}
   \vspace{-1mm}
      \caption{Recall@64 differences between BLINK and MuVER on entities with 1 to 100 sentences in their descriptions. We partition the entities by the number of sentences in entity descriptions and calculate metrics within each bin. The size for each bin is 5. 
      } \label{fig:effect_length}
   \vspace{-2mm}
\end{figure}

\paragraph{Effect on entities with long descriptions}
 As shown in Figure \ref{fig:effect_length}, existing EL systems (like BLINK) obtain passable performance on entities with short descriptions but fail to manage those well-populated entities as the length of descriptions increases. For instance, the error rate of BLINK is 7.79\% for entities with 5-10 sentences but 39.91\% for entities with 75-80 sentences, which is more likely to contain various aspects for the entity. MuVER demonstrates its superiority over entities with long descriptions, which significantly reduces the error rate to 17.65\% (-22.06\%) for entities with 75-80 sentences while maintains the performance on entities with short descriptions, which achieves the error rate of 6.78\% (-1.01\%) for entities with 5-10 sentences.  

\subsection{KB: Wikipedia}
We test AIDA-B, MSNBC and WNED-WIKI on the version of Wikipedia dump provided in KILT \citep{kilt2020}, which contains 5.9M entities. Implementation details are listed in Appendix \ref{apx:AIDA_implementation}. BLINK performance on these datasets is reported in its official Github repository\footnote{\url{https://github.com/facebookresearch/BLINK}}. 
We report the In-KB accuracy in Table \ref{tbl:wiki} and observe that MuVER out-performs BLINK on all datasets except the recall@100 on WNED-WIKI. 

\section{Related Work}
Representing each entity with a fixed-sized vector has been a common approach in Entity Linking. \citet{Ganea2017Deep} defines a word-entity conditional distribution and samples positive words from it. The representations of those positive words aim to approximate the entity embeddings compared with random words. \citet{Yamada2016Joint} models the relatedness between entities into entity representations. NTEE \citep{YamadaSTT17} trains entity representations by predicting the relevant entities for a given context in DBPedia abstract corpus. \citet{ReLICLing} and \citet{Yamada20} pre-train variants of the transformer-based model by maximizing the consistency between the context of the mentions and the corresponding entities. Those entity representations suffer from a cold-start problem that they cannot link mentions to unseen entities. 

Another line of work is to generate entity representations using entity textual information, such as entity descriptions. \citet{LogeswaranCLTDL19} introduces an EL dataset in the zero-shot scenario to place more emphasis on reading entity descriptions. BLINK \citep{WuPJRZ20} proposes a bi-encoder to encode the descriptions and enhance the bi-encoder by distilling the knowledge from the cross-encoder. \citet{YaoCP20} repeats the position embedding to solve the long-range modeling problem in entity descriptions.
\citet{Zhang2021under} demonstrates that hard negatives can enhance the contrast when training an EL model. 

\section{Conclusion}
In this work, we propose a novel approach to construct multi-view representations from descriptions, which shows promising results on four EL datasets. Extensive results demonstrate the effectiveness of multi-view representations and the heuristic search strategy. In the future, we will explore more reliable and efficient approaches to construct views.

\section*{Acknowledgement}
 This work was supported by Alibaba Group through Alibaba Innovative Research Program and was also supported by the Key Research and Development Program of Zhejiang Province, China (No. 2021C01013), the National Key Research and Development Project of China (No. 2018AAA0101900), the Chinese Knowledge Center of Engineering Science and Technology (CKCEST) and MOE Engineering Research Center of Digital Library. We thank Gongfan Fang, Yongliang Shen for their valuable comments on writing this paper.

\bibliography{anthology,custom}
\bibliographystyle{acl_natbib}

\clearpage

\appendix
\section{Appendix}
\label{sec:appendix}

\begin{table*}[b]
   \centering
   \scalebox{0.85}{
   \begin{tabular}{c c c c c c c c c c}
      \midrule
      \midrule
      \bf{Domain} & \bf{Method}  & \bf R@1  & \bf R@2 & \bf R@4 & \bf R@8 & \bf R@16 & \bf R@32 & \bf R@50 & \bf R@64 \\
      \midrule
      \bf Forgotten & BLINK & 
      63.75 & 74.83 & 82.17 & 85.50 & 89.08 & 91.17 & 92.83 & 93.75  \\
      \bf Realms & MuVER  & 
      62.5 & 78.5 & 86.67 & 90.92 & 93.58 & 96.00 & 96.75 & 97.00  \\
      \midrule
      \multirow{2}{*}{\bf Lego}
      & BLINK  & 
      50.04 & 65.39 & 75.81 & 81.65 & 84.82 & 88.41 & 90.58 & 91.83 \\
      & MuVER  & 
      50.46 & 68.81 & 78.32 & 84.4 & 88.82 & 91.91 & 93.33 & 93.74  \\
      \midrule 
      \multirow{2}{*}{\bf Star Trek}
      & BLINK  & 
      49.28 & 60.07 & 68.87 & 74.26 & 78.94 & 82.47 & 84.62 & 85.88 \\
      & MuVER  & 
      47.95 & 62.17 & 71.28 & 77.45 & 82.40 & 86.87 & 89.19 & 90.32  \\
      \midrule 
      \multirow{2}{*}{\bf Yugioh}
      & BLINK  & 
      35.66 & 47.45 & 56.14 & 63.22 & 68.35 & 73.00 & 75.90 & 77.71 \\
      & MuVER  & 
      34.32 & 50.06 & 63.25 & 72.61 & 78.48 & 83.94 & 86.69 & 88.26  \\
      \midrule 
   \end{tabular} 
   }
   \caption{Recall@k on four unseen domains: Forgotten Realms, Lego, Star Trek and Yugioh. } \label{tbl:zeshel_domain}
\end{table*} 

\subsection{Statistics of datasets} \label{apx:statistic}
Table \ref{tbl:statistic} shows statistics for four entity linking datasets: AIDA, MSNBC, WNED-WIKI and ZESHEL. MSNBC and WNED-WIKI are two out-of-domain test sets, which are evaluated on the model trained or finetuned on AIDA-train.
\begin{table}[h]
   \centering
   \scalebox{0.75}{
   \begin{tabular}{c |c |c |c |c}
      \midrule
      \midrule
       \multicolumn{2}{c|}{Dataset} & Mention Num & KB & Entity Num\\
       \midrule
       \multirow{3}{*}{AIDA} & Train & 18448 & \multirow{5}{*}{\shortstack{Wiki-\\pedia}}& \multirow{5}{*}{5903530} \\
        & Valid(A) & 4791 & & \\
        & Test(B) & 4485 & & \\
       \cmidrule{1-3}
       \multicolumn{2}{c|}{MSNBC} & 656 & & \\
       \multicolumn{2}{c|}{WNED-WIKI} &6821 &  &\\
      \midrule
      \multirow{3}{*}{ZESHEL}
      & Train & 49275 & \multirow{3}{*}{Wikia} & 332632\\
      & Valid & 10000 & & 89549 \\
      & Test & 10000 & & 70140\\
      \midrule
   \end{tabular} 
   }
   \caption{Statistics of four EL datasets.} \label{tbl:statistic}
\end{table} 

\subsection{Implementation Details} \label{apx:AIDA_implementation}
\paragraph{ZESHEL} We have reported the best-performing hyperparameter in Section 3.2. Here we show the search bounds for the hyperparameters. We perform grid search on learning rate, weight decay, warmup ratio and batch size:
\begin{itemize}
    \item Learning rate: [$5\text{e}^{-6}$, $1\text{e}^{-5}$, $2\text{e}^{-5}$, $5\text{e}^{-5}$]
    \item Weight decay: [0.1, 0.01, 0.001]
    \item Warmup ratio: [0, 0.1]
    \item Batch size: [32, 64, 128, 196]
\end{itemize}

\paragraph{AIDA} We finetune MuVER based on the EL model released by BLINK, which is pretrained on 9M annotated mention-entity pairs. Unlike the experiments on ZESHEL that adopting in-batch random negatives to train our model, we add hard negatives in batch. Due to the vast size of entities in Wikipedia, randomly sampled negatives are always too simple for the model to extract semantic features, thus degrading performance. We finetune our model on AIDA-CoNLL train set for one epoch. Batch size is set to 8. We add 3 hard negatives for each mention into the random in-batch negatives, which are precomputed using BLINK. The number of views is 5 for each entity and we choose the first 5 paragraphs with first 40 tokens, which are more likely to be summarizations. Other hyperparameters are consistent with configurations on ZESHEL. 

\paragraph{Parameters for MuVER} Since MuVER has two BERT encoders, it has twice the number of parameters as BERT, which are listed in Table \ref{tbl:parameters}.

\begin{table}[h]
   \centering
   \begin{tabular}{c c}
      \midrule
      \midrule
      Model & Number of parameters\\
      \midrule
      MuVER (base) & 220M \\
      MuVER (large) & 680M \\
      \midrule
   \end{tabular} 
   \caption{Numbers of parameters for MuVER. MuVER (base) is used in ZESHEL and MuVER (large) is used in datasets under full Wikipedia setting.} \label{tbl:parameters}
\end{table}

\subsection{Performance on Unseen Domains} \label{apx:unseen}
In Table \ref{tbl:zeshel_domain}, we compare MuVER with BLINK on four unseen domains on ZESHEL. We observe a significant improvement on all four unseen domains, especially on Yugioh, which achieves +11.35 points on Recall@64. Furthermore, MuVER can reach comparative performance with BLINK's top-64 candidates by retrieving around 16-32 candidates, which reduces the computational cost for entity ranking.



\end{document}